\documentclass[12pt]{iopart}

\usepackage[english]{babel}
\usepackage{multirow}
\usepackage{graphicx}
\usepackage{array}
\usepackage{subfigure}
\usepackage{multirow}
\usepackage{graphicx}
\usepackage{array}
\usepackage{subfigure}

\usepackage{natbib}
\usepackage{iopams}

\usepackage{hyperref}

\begin{document}

\title[]{Phase function estimation from a diffuse optical image via deep learning}

\author{Yuxuan Liang$^1$, Chuang Niu$^2$, Chen Wei$^3$, Shenghan Ren$^3$, Wenxiang Cong$^2$ and Ge Wang$^2$\footnote{Author to whom any correspondence should be addressed.}}
\address{$^1$ School of Physical Sciences, University of Science and Technology of China, Hefei, Anhui 230026, China}
\address{$^2$ Biomedical Imaging Center, Center for Biotechnology \& Interdisciplinary Studies, Department of Biomedical Engineering, Rensselaer Polytechnic Institute, Troy, NY 12180, USA}
\address{$^3$ School of Electronic Engineering, Xidian University, Xi'an, Shaanxi 710071, China}
\ead{\mailto{wangg6@rpi.edu (Ge Wang)}}
\vspace{10pt}

\begin{abstract}
The phase function is a key element of a  light propagation model for Monte Carlo (MC) simulation, which is usually fitted with an analytic function with associated parameters. In recent years, machine learning methods were reported to estimate the parameters of the phase function of a particular form such as the Henyey-Greenstein phase function but, to our knowledge, no studies have been performed to determine the form of the phase function. Here we design a convolutional neural network to  estimate the phase function from a diffuse optical image without any explicit assumption on the form of the phase function. Specifically, we use a Gaussian mixture model as an example to represent the phase function generally and learn the model parameters accurately. The Gaussian mixture model is selected because it provides the analytic expression of phase function to facilitate deflection angle sampling in MC simulation, and does not significantly increase the number of free parameters. Our proposed method is validated on MC-simulated reflectance images of typical biological tissues using the Henyey-Greenstein phase function with different anisotropy factors. The effects of field of view (FOV) and spatial resolution on the  errors are analyzed to optimize the estimation method. The mean squared error of the phase function is 0.01 and the relative error of the anisotropy factor is 3.28\%. 

\end{abstract}

\vspace{2pc}

\section{Introduction}

Optical properties of biological tissues are important in developing optical technologies for diagnostic and therapeutic applications. The commonly used optical properties include the absorption coefficient ($u_a$), reduced scattering coefficient ($\mu_s'$), scattering coefficient ($u_s$), scattering phase function ($p(\theta, \psi)$ where $\theta$ and $\psi$ are the deflection and azimuthal angles respectively, scattering anisotropy ($g$), and tissue refractive index ($n$) \citep{2013_PMB_Overview}. Since optical parameters are directly related to the structural and biochemical properties of tissues, they  provide contextual information and reflect physiological and pathological states \citep{2014_JBO_Influence}. 

Generally speaking, techniques for measuring tissue optical properties can be divided into direct and indirect. Each type of methods has its advantages and limitations \citep{1995_Wilson_book}. The direct methods are limited to optically thin samples that typically require complicate preparation procedures. The indirect methods estimate the optical parameters by solving the inverse problem of a light propagation model. Since it is not limited to thin samples, indirect methods suit well nondestructive applications; for example, evaluation of kidney transplant viability \citep{2019_JBO_Kidney} and tissue damage \citep{2021_SR}. The indirect measurement methods involve either analytic approximations of the radiative transfer model \citep{1978_Book_RTE} (such as diffusion approximation equation \citep{2007_Wang_BOBook}, simplified spherical harmonics equations \citep{2006_JCP_SPN}, and hybrid models \citep{2016_TBME_Chen}) or Monte Carlo (MC) simulation \citep{1995_Wang_MCML}. Both types of methods usually adopt an iterative procedure. Despite the huge computational burden, MC simulation is very successful in many applications \citep{1999_JBO_GK}, and has become the gold standard for studying the light propagation in biological tissues \citep{2017_RBME_MC_Review}. 

When performing the MC simulation, a prerequisite is the knowledge of the phase function, which describes the probability distribution for light to be deflected into different scattering angles at each scattering location \citep{2015_JBO_Detecting}. Since direct measurement of the phase function using a goniophotometer is limited to thin samples and is particularly difficult for media with high anisotropy factor \citep{1999_JBO_GK}, the analytically approximated phase functions are appealing in biomedical studies. Many such phase functions were formulated in the literature, such as the Henyey-Greenstein \citep{1941_HG}, modified Henyey-Greenstein \citep{1999_JOSA_MHG} and Gegenbauer kernel \citep{1999_JBO_GK}. Among them, the heuristic HG function, although proposed 80 years ago, is the most commonly used in the MC software \citep{1995_Wang_MCML, 2009_OE_Fang, 2013_Ren_MOSE}. However, the validity and accuracy of these phase functions depend heavily on tissue characteristics and experimental conditions, as demonstrated in \citep{2019_SPIE_FromMC}. Due to the optically heterogeneous nature of biological tissues, the selection of the phase function and the estimation of its parameters remains a rather challenging task, especially when the tissue geometry is complex.

In the past few years, the rapid advancement in artificial intelligence, especially deep learning, has revolutionized many areas of research such as computer vision and natural language processing \citep{2021_ACM_3Big}, tomographic imaging \citep{WangGe2020Book}, and photonics \citep{2020_AIPhotonics}. Specifically, for tissue optical parameter estimation, a neural network was proposed to estimate $\mu_a$, $\mu_s'$ and two subdiffusive parameters ($\gamma$ and $\delta$) from subdiffusive spatially resolved reflectance \cite{2018_OL_Efficient}, where the simulated observations were directly fed into a regression network consisting of three fully connected (FC) layers. Similarly, FC networks were also employed in \cite{2018_OL_Deep} and \cite{2021_Sun} but with more layers to estimate $\mu_a$ and $\mu_s'$ using simulated data in the frequency spatial domain. In a similar spirit, a four-layer FC network was designed \cite{2021_SR} to estimate $\mu_a$, $\mu_s'$ and $n$ simultaneously in terms of the moments of the  profiles of spatial-temporal diffuse reflectance. 

Although the aforementioned studies indeed leveraged deep learning methods, none of them have explored the feasibility of estimating the functional form of the phase function. Motivated by the observation that the analytic approximations of the phase function with a few free parameters may not be sufficiently accurate to quantify the {\it in vivo} scattering characteristics in biomedical studies, in this paper we propose to estimate the functional form of the phase function directly from diffuse reflectance data through deep learning. For this purpose, we design a dedicated convolutional neural network (CNN), and conduct extensive simulations on 11 typical biological tissues using a modified graphics processing units (GPU)-accelerated MC software package CUDAMCML \citep{2008_CudaMCML}, which is based on the pioneering work on the steady-state MC simulation of multi-layered turbid media, which is widely known as MCML \citep{1995_Wang_MCML}. Whereas the experiments were performed in the continuous wave (CW) mode, the proposed method can in principle be easily extended to other imaging modes (such as spatial-temporal and spatial-frequency).

Our main contributions can be summarized as follows. First, We propose the first data-driven CNN-based inverse MC model to estimate the scattering phase function from diffuse reflectance data. The functional form of the phase function is assumed to be in the space of the Gaussian mixture model (GMM), which is quite general and computationally efficient (the analytic expression of the phase function facilitates deflection angle sampling and does not significantly increase the number of free parameters). Also, we investigate the effects of field of view (FOV) and spatial resolution on the accuracy of the phase function estimation and optimize the estimation accuracy, which promises a wide array of applications in the field of biomedical optics.

\section{Materials and Methods}


\subsection{Biological Tissues}

This study focuses on the estimation of the functional form of the phase function in the MC simulations with 11 typical biological tissues of known $\mu_a$ and $\mu_s'$ and varying $g$ values. This choice is made based on the fact that existing techniques are more robust for measurement of $\mu_a$ and $\mu_s'$ but not so for $\mu_s$ and $g$ due to the diffusive nature of biological tissues \citep{2013_PMB_Overview}. The $\mu_a$ and $\mu_s'$ values of the selected tissues are calculated based on the empirical functions with the parameters provided in \cite{Alexandrakis_2005}. The photon wavelength is 670 nm, at which the absorption and reduced scattering coefficients are listed in Table \ref{tab_tissueParam}. Based on the classification criteria and empirical thresholds summarized in \cite{2016_TBME_Chen}, the tissues are classified into high scattering and low scattering categories. Among the 11 tissues, liver, spleen, muscle, and whole blood are low scattering tissues, while the others are high scattering tissues. 

There are several MC software packages available, including MCML \citep{1995_Wang_MCML}, MCX \citep{2009_OE_Fang} and MOSE \citep{2013_Ren_MOSE} for optical transport simulation. The absorption coefficient $\mu_a$, scattering coefficient $\mu_s$, and anisotropy factor $g$ are essential for MC simulation. The reduced scattering coefficient incorporates the scattering coefficient $\mu_s$ and the anisotropy $g$: $\mu_s'=\mu_s(1-g)$. The $g$ value is selected in Section \ref{section_datasets}. The use of $g$ implies that the phase function is the Henyey-Greenstein function, which is the only type of the phase function supported by CUDAMCML and most other existing MC software packages. However, this is not a problem for validating our proposed neural network for phase function estimation.

\small
\begin{table}
\caption{\label{tab_tissueParam} Absorption and reduced scattering coefficients of the tissues.} 
\begin{indented}
\lineup
\item[]\begin{tabular}{@{}*{4}{l}}
\br                              
Tissue & $\mu_a$ (cm$^{-1}$) & $\mu_s'$ (cm$^{-1}$)  & Scattering category\\ 
\mr
Adipose         & 0.038  & 12.077 & high  \\
Bone            & 0.603  & 24.953 & high  \\
Bowel           & 0.117  & 11.490 & high  \\
Heart wall      & 0.583  & 9.639  & high  \\
Kidneys         & 0.654  & 22.530 & high  \\
Liver and spleen& 3.490  & 6.781  & low   \\
Lung            & 1.948  & 21.739 & high  \\
Muscle          & 0.863  & 4.291  & low   \\
Skin            & 0.699  & 22.190 & high  \\
Stomach wall    & 0.113  & 14.369 & high  \\
Whole blood     & 11.621 & 18.140 & low   \\
\br
\end{tabular}
\end{indented}
\end{table}\normalsize

\subsection{MC Simulation}

Diffuse reflectance images of tissues are simulated using CUDAMCML \citep{2008_CudaMCML}, a GPU-accelerated implementation of MCML \citep{1995_Wang_MCML}. Since MCML uses a Cartesian coordinate system to trace photon packets and a cylindrical coordinate system to record  diffuse reflectance signals, we modified CUDAMCML to record the Cartesian coordinates of photon packets as they hit the tissue surface to generate reflectance images in the xy-plane. The spatial resolution $\Delta r$ in the radial direction defines the resolution in the xy-plane. The size of the reflectance image $W\times H$ (in pixels) is determined by the total number of grid elements $N_r$ in the radial direction, i.e., $W=H=2 N_r +1$. For simplicity, the FOV of the reflectance image is reported as $2 N_r \Delta r \times 2 N_r \Delta r$. The modified CUDAMCML is build with CUDA 11.1 under Xubuntu 18.04. 

\subsection{Phase Function Representation}

Estimating the phase function is basically a regression problem, which can be implemented using a neural network as follows:
\begin{equation}
  p(\theta, \psi) = f (I_R(\mu_a, \mu_s, g)),
\end{equation}
where $I_R$ is a reflectance image generated in the MC simulation and $f$ is the nonlinear feedforward mapping of the neural network. In thicker tissues where multiple scattering occurs, such as the semi-infinite tissues used in this study, it is valid to express the scattering phase function $p(\theta, \psi)$ as a function of the polar angle $\theta$ while omitting the dependence on the azimuthal angle $\psi$ \citep{2013_PMB_Overview,2016_JBO_Quantifying}. 

The popular Henyey-Greenstein function is well known as 
\begin{equation}
  \label{eq_HG}
  p(\theta) = \frac{1}{4\pi}\frac{1-g^2}{(1+g^2-2g\cos(\theta))^{3/2}},
\end{equation}
where the anisotropy factor $g$ is defined as the average cosine of the scattering angle, i.e., $g=\langle cos(\theta) \rangle$. Since the integral of $\tilde p(\theta)=2\pi p(\theta) \sin(\theta)$ over $\pi$ is unity, the normalized Henyey-Greenstein  function $\tilde p(\theta)$ is used in the following. Without confusion, it is still denoted as $p(\theta)$. In MC implementations, Equation \ref{eq_HG} is often written as a function of $\mu=cos(\theta)$:
\begin{equation}
  \label{eq_HG2}
  p(\mu) = \frac{1}{2}\frac{1-g^2}{(1+g^2-2g\mu)^{3/2}}, 
\end{equation}
which facilitates the sampling of the deflection angle at each scattering location based on the analytic inverse CDF of $p(\mu)$ \citep{1995_Wang_MCML}.

Unlike the other methods that estimate the unknown parameters of a known function form of the phase function, the goal of our study is to use a neural network to estimate both the form and parameters of the phase function based on reflectance signals. An intuitive approach is to use $p(\theta)$ or $p(\mu)$ as the target for the neural network to learn from training data. However, this leads to two problems. The first is how to discretize $\theta$ or $\mu$, which determines the number of neurons in the output layer of the neural network. Too few grids would lead to too large fitting errors but too many grids would take too much computational resources. The other problem is that when the discretized phase function representation is used in the MC simulation, numerical sampling methods, such as the lookup table-based methods \citep{2017_BOE_LUT}, are required but these methods are rather memory intensive. 

Considering the above two issues, we propose to employ an analytic representation, the widely used Gaussian mixture model (GMM), to specify the functional form of the underlying phase function. A GMM is defined as the superposition of some basic Gaussian densities in the following form
\begin{equation}
  \label{eq_GMM}
  p_{GMM}(x) = \sum_{k=1}^{K} \pi_k \mathcal{N} (x|m_k, \sigma_k),
\end{equation}
where $\pi_k$, $m_k$ and $\sigma_k$ are the mixing coefficient, mean and variance of the $k^{th}$ Gaussian density $\mathcal{N} (x|m_k, \sigma_k)$, respectively, and $K$ is the number of Gaussian components (NoG). The mixing coefficients must satisfy the requirements of $0 \leq \pi_k \leq 1$ and $\sum_{k=1}^{K} \pi_k = 1$. With a sufficient number of Gaussian components and suitable mixing coefficients, GMM can approximate almost any continuous density up to an arbitrary accuracy \citep{2006_Bishop_PRML}.

\begin{figure}[!htbp]
  \begin{center}
    \subfigure[]{\includegraphics[scale=0.7]{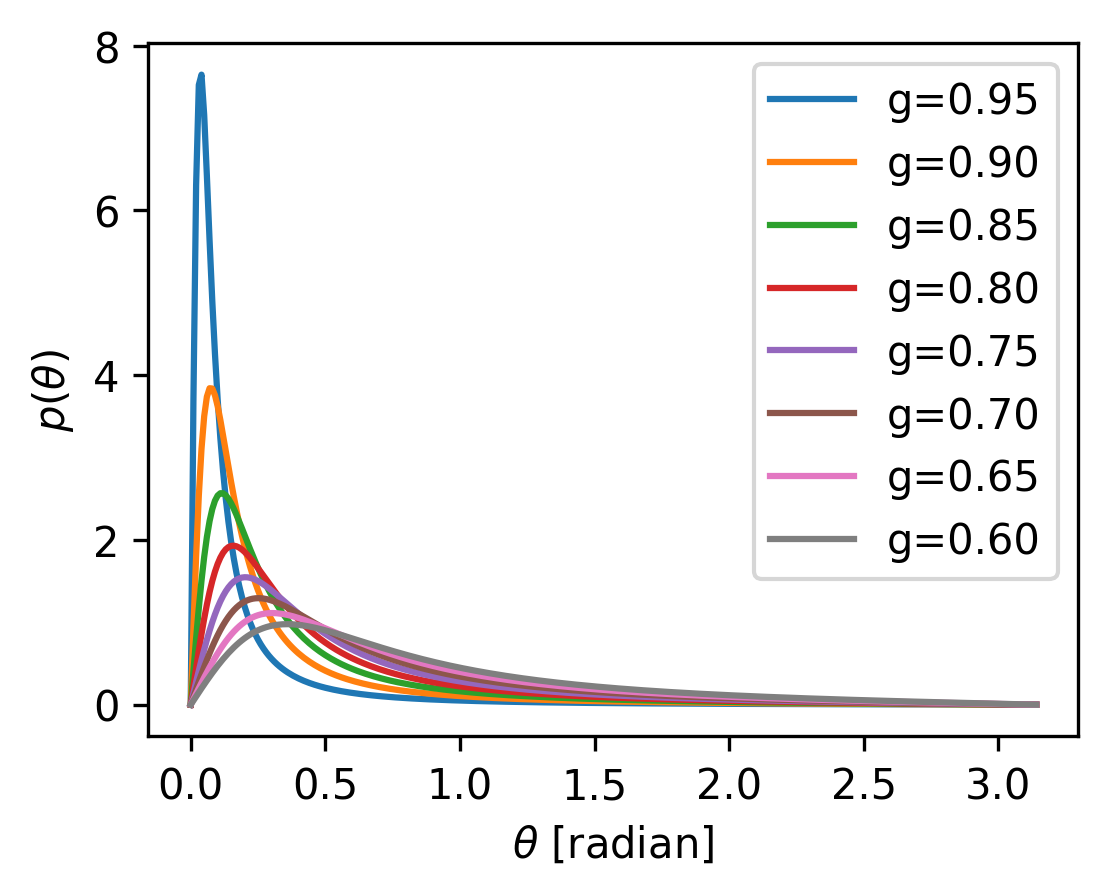}}
    \subfigure[]{\includegraphics[scale=0.7]{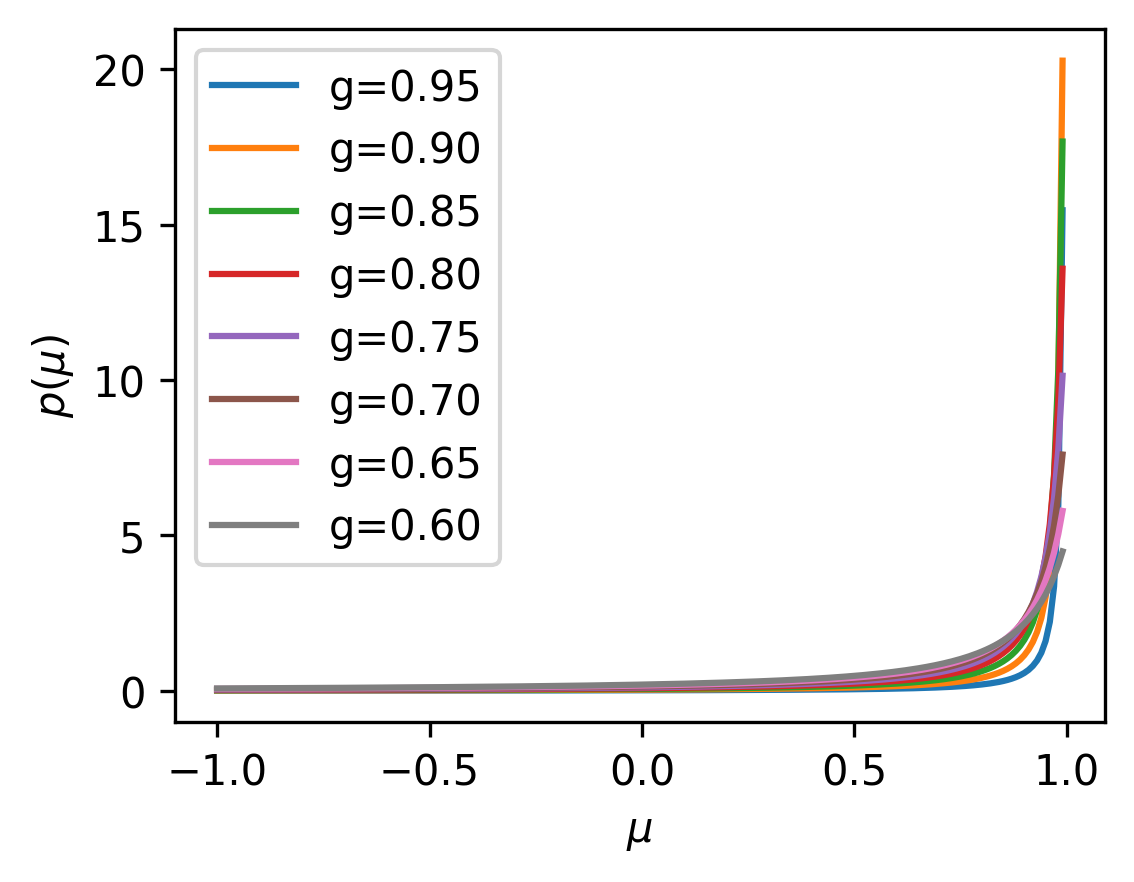}}
    \caption{Henyey-Greenstein functions with different anisotropy factors. (a) $\tilde p(\theta)$ and (b) $p(\mu)$ examples. }
    \label{fig_HG}
  \end{center}
\end{figure}

In the phase function estimation, there are two choices for the argument in Equation \ref{eq_GMM}: $\theta$ or $\mu=\cos(\theta)$. Figure \ref{fig_HG} plots $p(\theta)$ and $p(\mu)$ for representative $g$ values. Due to the symmetry of the Gaussian density, the asymmetric form of $p(\mu)$ makes it difficult to be approximated by GMM. Hence, we choose the normalized Henyey-Greenstein function $p(\theta)$ as the target for the neural network to learn. It can be seen in Figure \ref{fig_HG}(a) that the larger the anisotropy factor $g$, the narrower and steeper the Henyey-Greenstein function. Therefore, for large $g$ values, it is more difficult to fit the Henyey-Greenstein function using GMM, which is also confirmed in the simulation experiments below.   

\subsection{Neural Network Model}
\label{section_NN}

 To design a neural network model for our purpose, several factors should be considered as follows:

{\it Network architecture:} In this feasibility study, we adopt the ResNet-18 model \citep{2016_CVPR_ResNet} as the network architecture. The original ResNet-18 model is modified in the following aspects: (1) The channel number of the first layer is changed to 1 to be consistent with the single-channel reflection images; (2) The number of neurons in the last FC layer is set to 3K, corresponding to $\pi_k$, $m_k$ and $\sigma_k$ for the $K$ Gaussian components; and (3) A sigmoid layer is appended after the FC layer to normalize the outputs to the range [0, 1]. The modified network model is referred to as PhaseNet. 

{\it Loss function:} The mean squared error (MSE) is adopted to train and test PhaseNet. To calculate the loss function, $\theta$ is discretized into 1,000 grids. Because $p(\theta)$ is defined in the interval $[0,\pi]$ but $p_{GMM}(\theta)$ is not, $p_{GMM}(\theta)$ is truncated to $[0,\pi]$ and then normalized for calculation of the MSE. From $p_{GMM}(\theta)$, the anisotropy factor $g$ can be estimated as  

\begin{equation}
  \label{eq_g_estimation}
  \hat{g} = \int_{\theta=0}^{\pi} p_{GMM}(\theta) \cos(\theta).
\end{equation}

{\it Model selection:} Since the backbone of PhaseNet is fixed as ResNet-18, the NoG becomes the key parameter to determine the PhaseNet architecture. The optimal NoG for a given training dataset can be determined via cross-validation or with an analytic criterion such as the Akaike information criterion (AIC) or Bayesian information criterion (BIC) \citep{2006_Bishop_PRML}. In our experiments, it is found that AIC and BIC tend to favor overly simple models. This is because AIC and BIC use a term penalizing the number of model parameters. In PhaseNet, the number of free parameters in the FC layer is $512 \times (3 \times \rm{NoG} + 1)$, where 512 is the number of output channels of the global pooling layer in ResNet-18. Therefore, an increment of NoG leads to a significant boost in the number of the FC layer parameters, which is strongly discouraged by AIC and BIC. Therefore, we use cross-validation for model selection. 

The commonly used $k$-fold cross-validation scheme randomly divides a dataset into $k$ mutually exclusive folds and then repeatedly selects one fold as the validation set and the remaining $k$-1 folds as the training set. This random sampling strategy is not effective in estimating the optimal NoG value, as demonstrated in our experiments, where the validation error has the same declining trend as the training error as the NoG increases. As a result, the maximal test NoG value is always recommended. We believe that this is caused by the similarity between the reflectance images. Although they are generated by running MC simulations independently, they are very similar when a large number of photons are used. Therefore, the validation set generated by randomly splitting the entire dataset cannot reveal well the generalizability of the model. Finally, we select a leave-one-$g$-out (LOGO) cross-validation method to determine the optimal NoG for PhaseNet. The LOGO cross-validation divides the entire dataset into training and validation sets according to the $g$ values of reflectance images. During each iteration of cross-validation, images with one $g$ value are kept as the validation set, while the remaining images with other $g$ values are used for training. 

\subsection{Experimental Setup}
\label{section_datasets}

Five reflectance image datasets were simulated using CUDAMCML to investigate the effects of FOV and spatial resolution on the estimation accuracy of the phase function. The details of the datasets are presented in Table \ref{tab_dataset}. All 11 tissue types were covered by each of the five datasets. Each dataset has its own training and test subsets. The diffuse reflectance images of the tissues were simulated separately using CUDAMCML in a semi-infinite slab geometry and an infinitely narrow incident beam with 10 million photons. The tissue thickness was set to 100 cm to ensure the validity of the semi-infinite assumption. 

In typical biological tissues, the anisotropy factor $g$ ranges between 0.7 and 0.99 \citep{1990_IEEE_Review}. In our experiments, the lower bound of $g$ was set to 0.6. For each tissue, $g$ varies from 0.65 to 0.95 with a step size of 0.1 to generate training data, and varies from 0.6 to 0.9 with a step size of 0.1 for the test data. For each  set of parameters ($\mu_a$, $\mu_s$ and $g$), 200 samples were generated for training, and 40 samples for testing. The total numbers of images for training and testing in each dataset are 8,800 and 1,760 respectively.

\small
\begin{table}
  \caption{\label{tab_dataset} Parameters of the reflectance image datasets.} 
  \begin{indented}
    \lineup
    \item[]\begin{tabular}{@{}*{6}{l}}
    \br                              
    Name & FOV (mm$^2$) & Resolution ($\Delta r$, mm) & $N_r$ & Image Size & NoG \\ 
    \mr
    $DS_1$    & 2$\times$2   & 0.02 & 50   & 99$\times$99   & 11$^{*}$  \\
    $DS_2$    & 4$\times$4   & 0.02 & 100  & 199$\times$199 & 11$^{\bigstar}$  \\
    $DS_3$    & 6$\times$6   & 0.02 & 150  & 299$\times$299 & 11$^{*}$  \\
    $DS_4$    & 4$\times$4   & 0.01 & 200  & 399$\times$399 & 11$^{*}$  \\
    $DS_5$    & 4$\times$4   & 0.04 & 50   & 99$\times$99   & 11$^{*}$  \\
    \br
    \end{tabular}\\
    $^{\bigstar}$ Determined through leave-one-$g$-out cross-validation. \\
    $^{*}$ The same value for $DS_2$.
  \end{indented}
\end{table}\normalsize

The optimal NoG value for each dataset can be determined through LOGO cross-validation. However, in order to compare the performance of PhaseNet on different datasets fairly, we keep all factors the same except for FOV and resolution. Therefore, the optimal NoG value of $DS_2$, which is associated with representative FOV and spatial resolution values, was uniformly applied to all other datasets. The optimal NoG for $DS_2$ is 11, determined by varying the NoG from 2 to 12. After obtaining the optimal NoG, the PhaseNet models were trained on each dataset separately. Due to the stochastic nature of the training process, five models were trained on each dataset respectively. 

\section{Experimental Results}
\label{section_results}

\subsection{Implementation Details}

The proposed PhaseNet model was implemented with Pytorch 1.8.2 on a single NVIDIA GeForce RTX 3090 graphics card. At the LOGO cross-validation and training stages, the models were trained with momentum SGD optimizer for 30 epochs with a momentum of 0.9 and a weight decay of 0.005. A step-wise learning rate decay scheduler was used with an initial learning rate of 0.005, step size of 10 epochs, and drop rate of 0.1. The batch size was 220 for all datasets. The commonly used horizontal and vertical flipping were utilized for data augmentation. At the testing stage, the five PhaseNet models for each dataset were respectively tested on the corresponding test data to produce the estimation errors (MSE) on average. The code of PhaseNet is publicly available at \url{https://github.com/liangyuxuan1/phasefunction2}.

\subsection{Representative Results}


\small
\begin{table}
  \caption{\label{tab_overview} Summary of the mean squared and relative errors.} 
  \begin{indented}
    \lineup
    \item[]\begin{tabular}{@{}*{4}{l}}
    \br                              
    Dataset (FOV) & MSE  & Gain (\%) & Relative Error of $g$ (\%)\\ 
    \mr
    $DS_1$ (2$\times$2@0.02)   & 0.012 $\pm$ 0.016   & -16.7 & 3.421 $\pm$ 4.426  \\
    $DS_2$ (4$\times$4@0.02)   & 0.010 $\pm$ 0.013   & --    & 3.280 $\pm$ 3.047  \\
    $DS_3$ (6$\times$6@0.02)   & 0.017 $\pm$ 0.013   & -41.2 & 3.910 $\pm$ 3.450  \\
    $DS_4$ (4$\times$4@0.01)   & 0.029 $\pm$ 0.022   & -65.5 & 5.232 $\pm$ 5.302  \\
    $DS_5$ (4$\times$4@0.04)   & 0.011 $\pm$ 0.016   & -9.1  & 3.151 $\pm$ 3.263  \\
    \br
    \end{tabular}
  \end{indented}
\end{table}\normalsize

Table \ref{tab_overview} summarizes our key experimental results. For simplicity, the units of FOV (mm$^2$) and spatial resolution (mm) are omitted, and an FOV is denoted together with the spatial resolution when necessary, e.g., 4$\times$4@0.02. The MSE is presented as the average over all tissues and anisotropy factors for each dataset. The performance gain is defined as the relative error with respect to the MSE of the reference dataset $DS_2$. The minus sign represents performance degradation. The relative error of $g$ is defined as $(|g-\hat{g}|/g)\times 100\%$.

As shown in Table \ref{tab_overview}, the best estimation accuracy was achieved on $DS_2$ (FOV=4$\times$4@0.02). Using the performance on $DS_2$ as the reference, the performance gain was computed for the other datasets. Reducing the FOV to 2$\times$2 ($DS_1$) or expanding it to $6\times$6 ($DS_3$) at the same resolution of 0.02 resulted in 16.7\% and 41.2\% drop in accuracy respectively. When keeping the FOV unchanged but increasing the resolution to 0.01 ($DS_4$) or decreasing it to 0.04 ($DS_5$), the estimation error was increased by 65.5\% and 9.1\% respectively. The relative errors of $g$ shared the trends of the MSE data. 


\begin{figure}[!htbp]
  \begin{center}
    \subfigure[]{\includegraphics[scale=0.55]{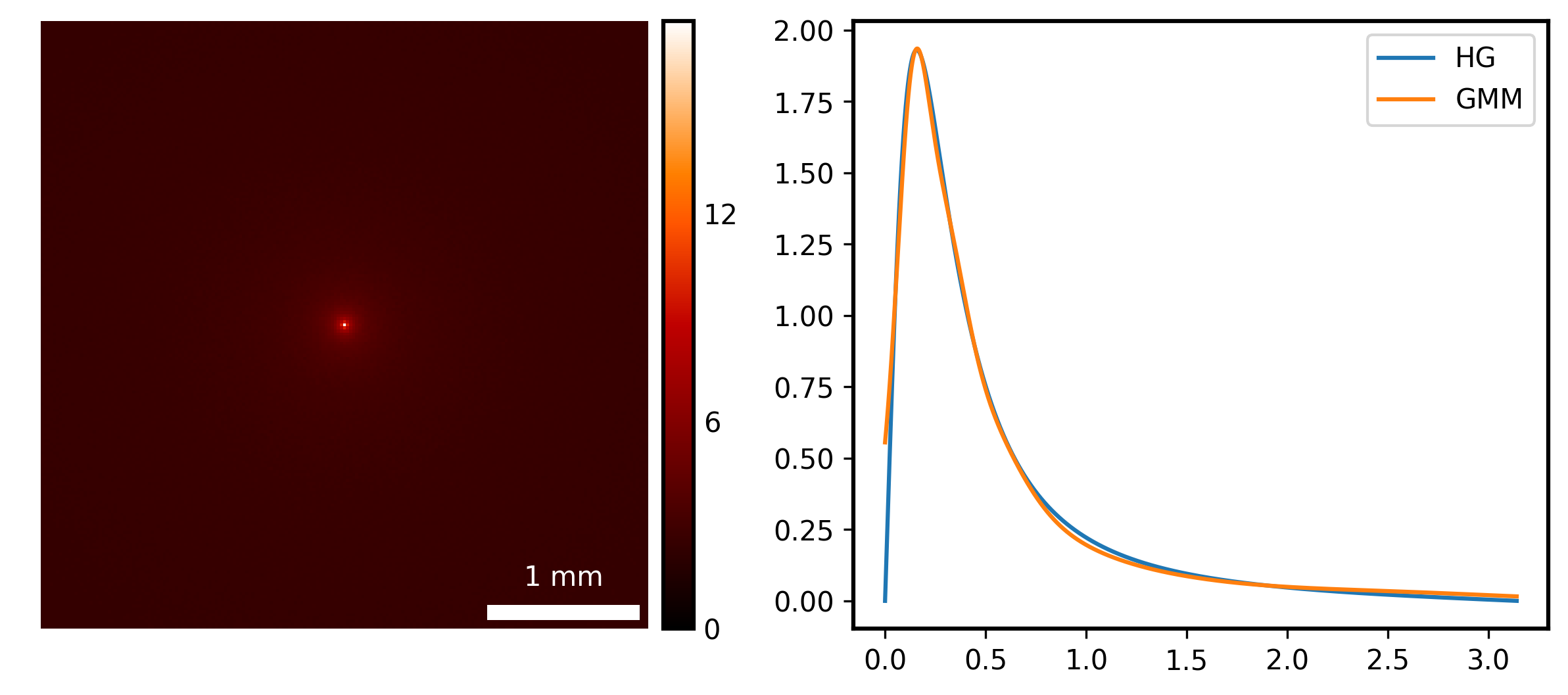}}
    \subfigure[]{\includegraphics[scale=0.55]{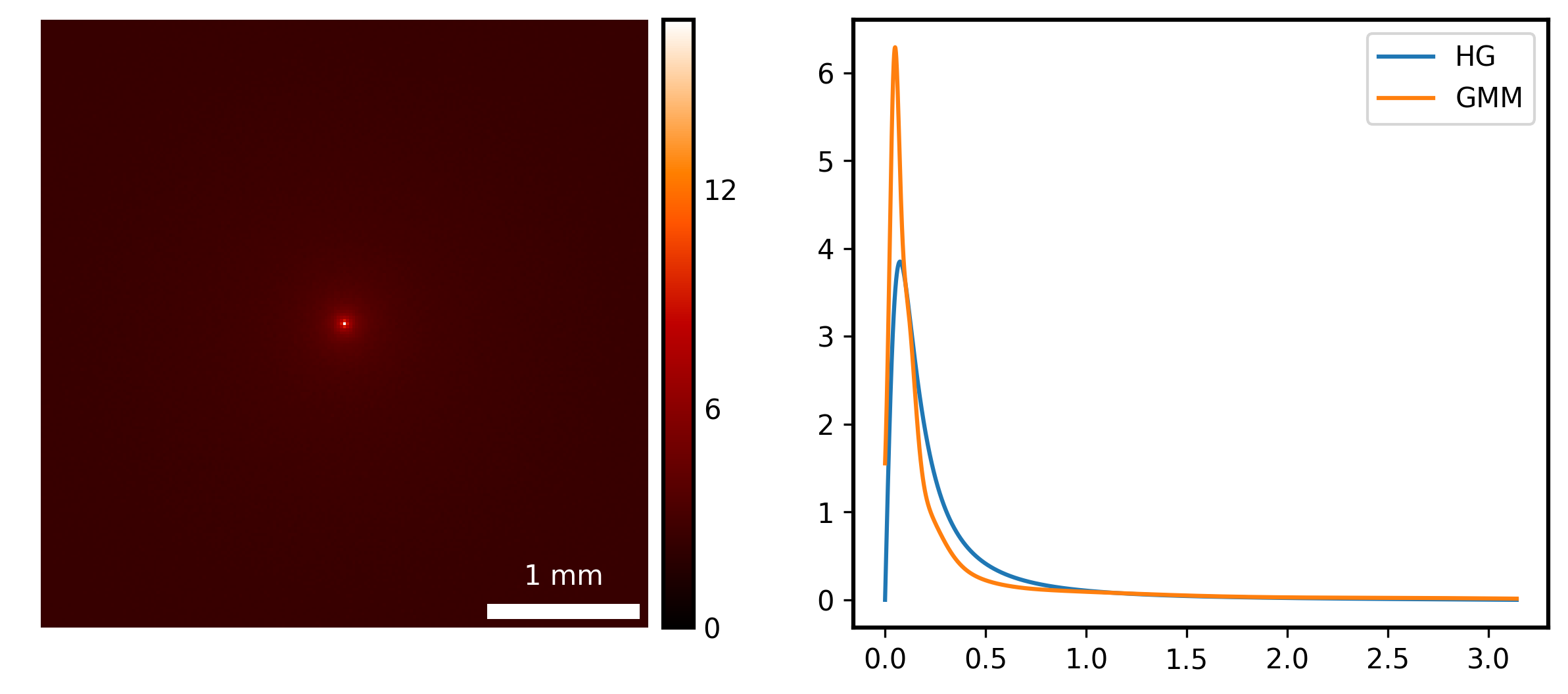}}
    \caption{Exemplary results. (a) An example of accurate estimation. Tissue: heart, $u_a=0.583$ cm$^{-1}$, $u_s=48.194$ cm$^{-1}$, $g=0.80$, MSE=0.002, and (b) an example of poor estimation. Tissue: heart, $u_a=0.583$ cm$^{-1}$, $u_s=96.387$ cm$^{-1}$, $g=0.90$, MSE=0.136. The dynamic range of the reflectance images on the left is compressed with the Gamma transformation ($\gamma=0.5$) for visualization. FOV=4$\times$4@0.02 is the same for both images. The ground truth and the estimated GMM are on the right. The x-axis of the phase function is $\theta$ in radian. }
    \label{fig_examples}
  \end{center}
\end{figure}

The MSE values do not provide sufficient information on how close the estimated phase functions is to the ground truth. However, the nonlinear nature of the phase function estimation (regression) prevents us from using the R-squared statistic to evaluate the fitness of PhaseNet. Therefore, we visually inspect the experimental results. Figure \ref{fig_examples} show two examples with the smallest and largest estimation errors on $DS_2$. In both examples, the reflectance images look very similar despite the different tissue parameters of $\mu_s$ and $g$. PhaseNet did capture the basic form of the Henyey-Greenstein phase function but apparently it was challenged near the boundary of $\theta=0$ and for large $g$ value such as 0.9, for which the phase function is much steeper and narrower. 

\subsection{Effect of FOV on Estimation Accuracy}

\begin{figure}[!htbp]
  \begin{center}
    \includegraphics[scale=0.6]{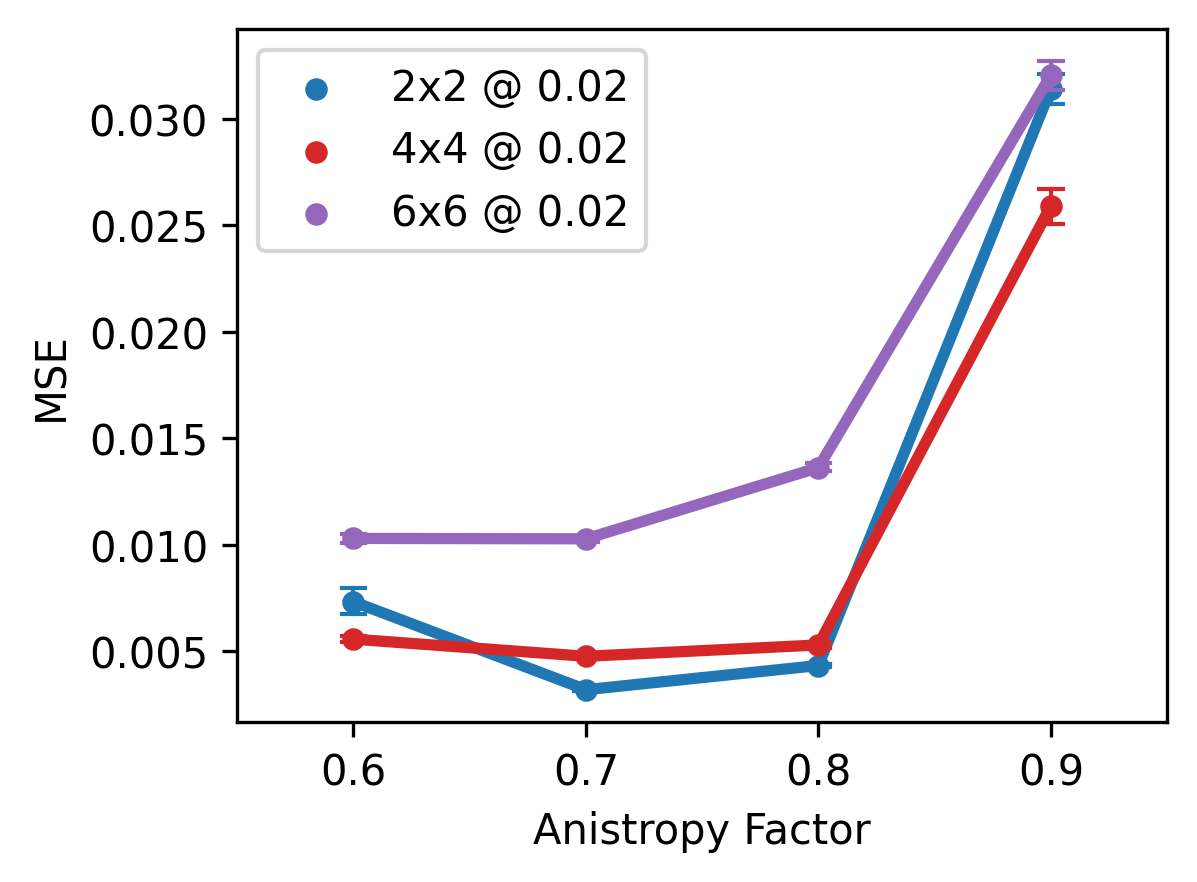}
    \caption{Phase function estimation errors at different FOVs, averaged across all the tissue types. }
    \label{fig_accuracy_FOV}
  \end{center}
\end{figure}

The MSE of the phase functions estimated by PhaseNet on $DS_1$, $DS_2$ and $DS_3$ are shown in Figure \ref{fig_accuracy_FOV}. These datasets have different FOVs but have the same spatial resolution of 0.02. At the first glance, the larger the FOV, the larger the estimation error. In order to quantify the statistical difference, the Wilcoxon rank sum test was used to test the differences in MSE measures across FOVs at each $g$ value separately because the errors are not normally distributed ($p<0.05$, Shapiro-Wilk test). The results indicate that the MSEs are significantly different ($p<0.05$) at all $g$ values except for that with FOV 2$\times$2 (blue dot) versus 6$\times$6 (purple dot) at $g$=0.9. The error plots for FOV 2$\times$2 and 4$\times$4 cross each other, but the average error of FOV 4$\times$4 is smaller. The estimation error of FOV 6$\times$6 is significantly larger than that with the other two FOVs.


\begin{figure}[!htbp]
  \begin{center}
    \includegraphics[scale=0.6]{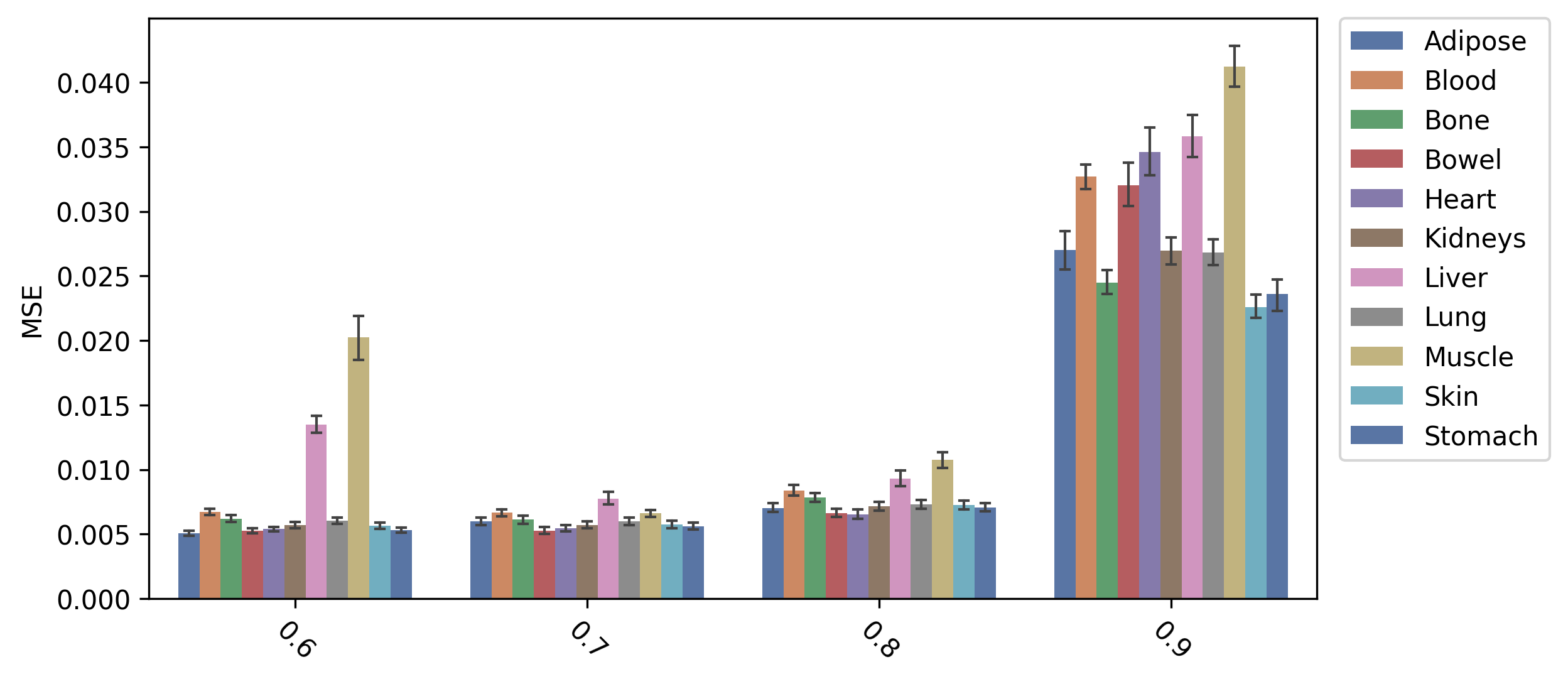}
    \caption{Phase function estimation errors for individual tissues, averaged across the FOVs in Figure \ref{fig_accuracy_FOV}. }
    \label{fig_tissue_accuracy_g_FOV}
  \end{center}
\end{figure}

Figure \ref{fig_tissue_accuracy_g_FOV} shows the estimation errors of individual tissues averaged across the FOVs. When $g=0.9$, the errors of all tissues are significantly larger compared to that at smaller  $g$ values. At $g$=0.6, the errors of muscle and liver are significantly larger than that of other tissues. According to Table \ref{tab_tissueParam}, these two tissues are low scattering tissues.



\subsection{Effect of Spatial Resolution on Estimation Accuracy}

\begin{figure}[!htbp]
  \begin{center}
    \includegraphics[scale=0.6]{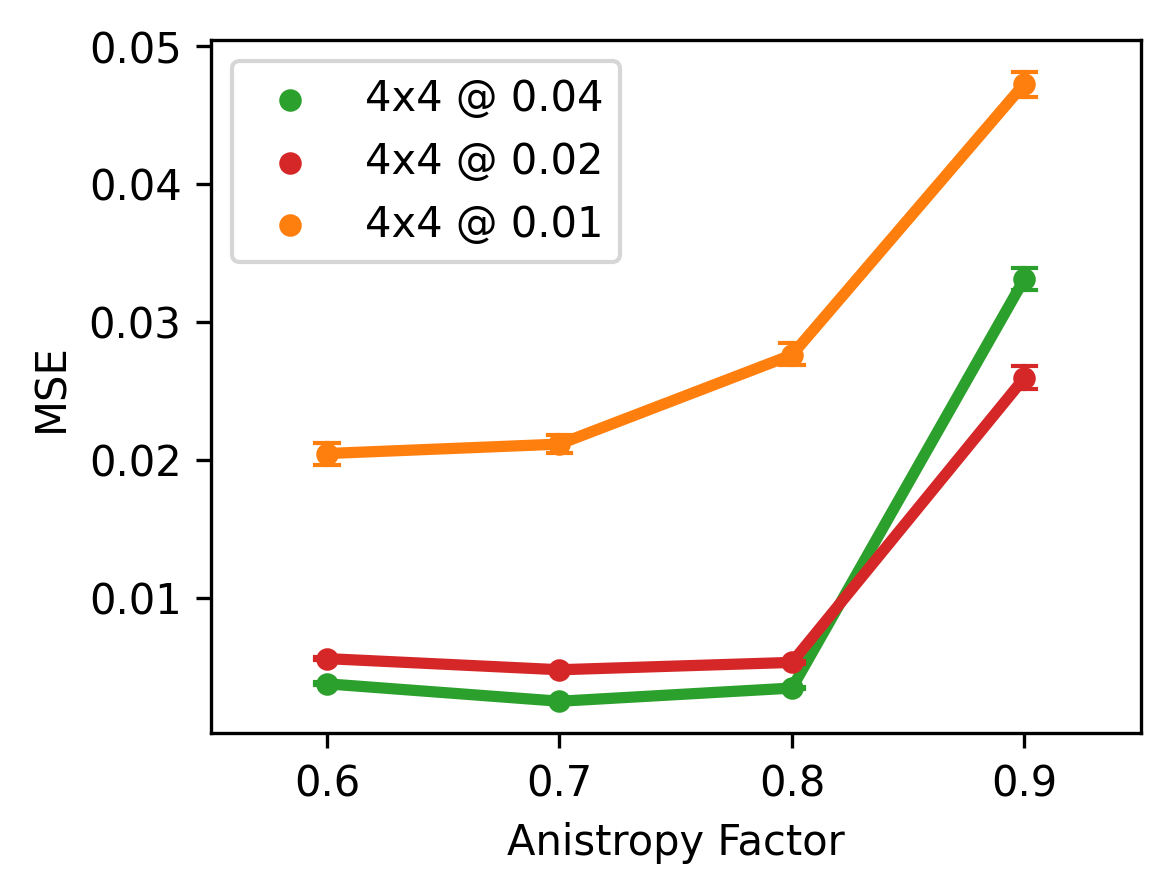}
    \caption{Phase function estimation errors at different spatial resolutions, averaged across all the tissues. }
    \label{fig_accuracy_Res}
  \end{center}
\end{figure}

Figure \ref{fig_accuracy_Res} shows the phase function estimation accuracy on $DS_2$, $DS_4$ and $DS_5$, where the FOV was fixed at 4$\times$4 while the spatial resolution varied from 0.02 ($DS_2$) to 0.01 ($DS_4$) or 0.04 ($DS_5$). Relative to the reference resolution of 0.02 (red dots), the estimation error was increased by 65.5\% (Table \ref{tab_overview}) when the resolution was increased to 0.01 (orange dots). When the resolution was reduced to 0.04 (green dots), the estimation error was increased by 9.1\%. The increment of the error is mainly due to the large error with $g$=0.9. At the other $g$ values, the estimation errors at resolution 0.04 were smaller than those at resolution 0.02. 

\begin{figure}[!htbp]
  \begin{center}
    \includegraphics[scale=0.6]{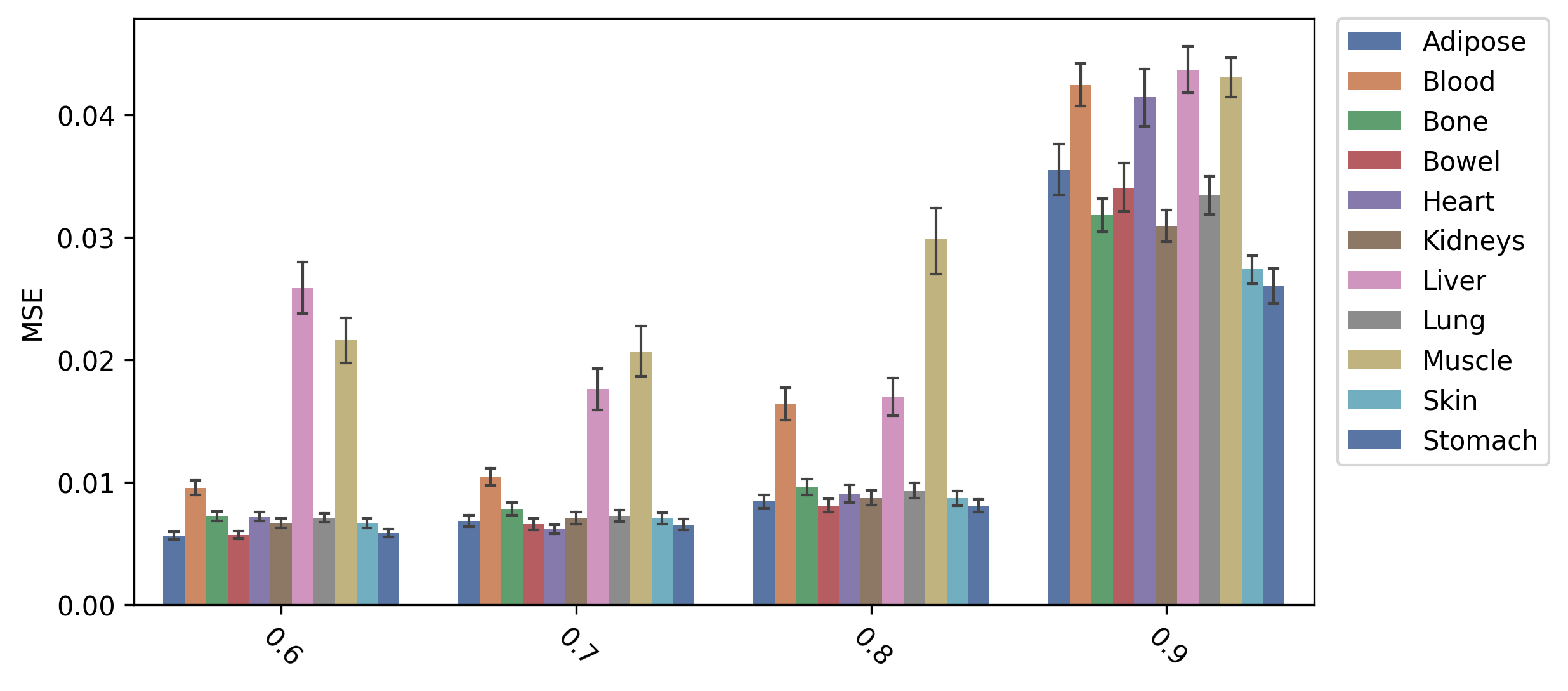}
    \caption{Phase function estimation errors for individual tissues, averaged across the resolutions in Figure \ref{fig_accuracy_Res}.}
    \label{fig_tissue_accuracy_g_Res}
  \end{center}
\end{figure}

Figure \ref{fig_tissue_accuracy_g_Res} shows the estimation errors for individual tissues, averaged across the different resolutions. Similar to the results in Figure \ref{fig_tissue_accuracy_g_FOV}, the errors are significantly larger at $g=0.9$ compared to that at smaller $g$ values. At the other $g$ values, the three low scattering tissues, muscle, liver, and blood, were subject to significantly higher errors than the high scattering tissues.

\section{Discussions}

Some of the above-described experimental results might appear somewhat unreasonable at the first glance. For example, increasing the FOV and improving the spatial resolution are unhelpful and even harmful in some cases to the estimation accuracy. In general, we always want an imaging system to have as high resolution as possible with a sufficiently large FOV to guarantee the measurement accuracy. To justify the results, we investigate the distribution of the reflectance signals and the similarity of the reflectance images for different tissues and anisotropy factors. 

\begin{figure}[!htbp]
  \begin{center}
    \subfigure[]{\includegraphics[scale=0.52]{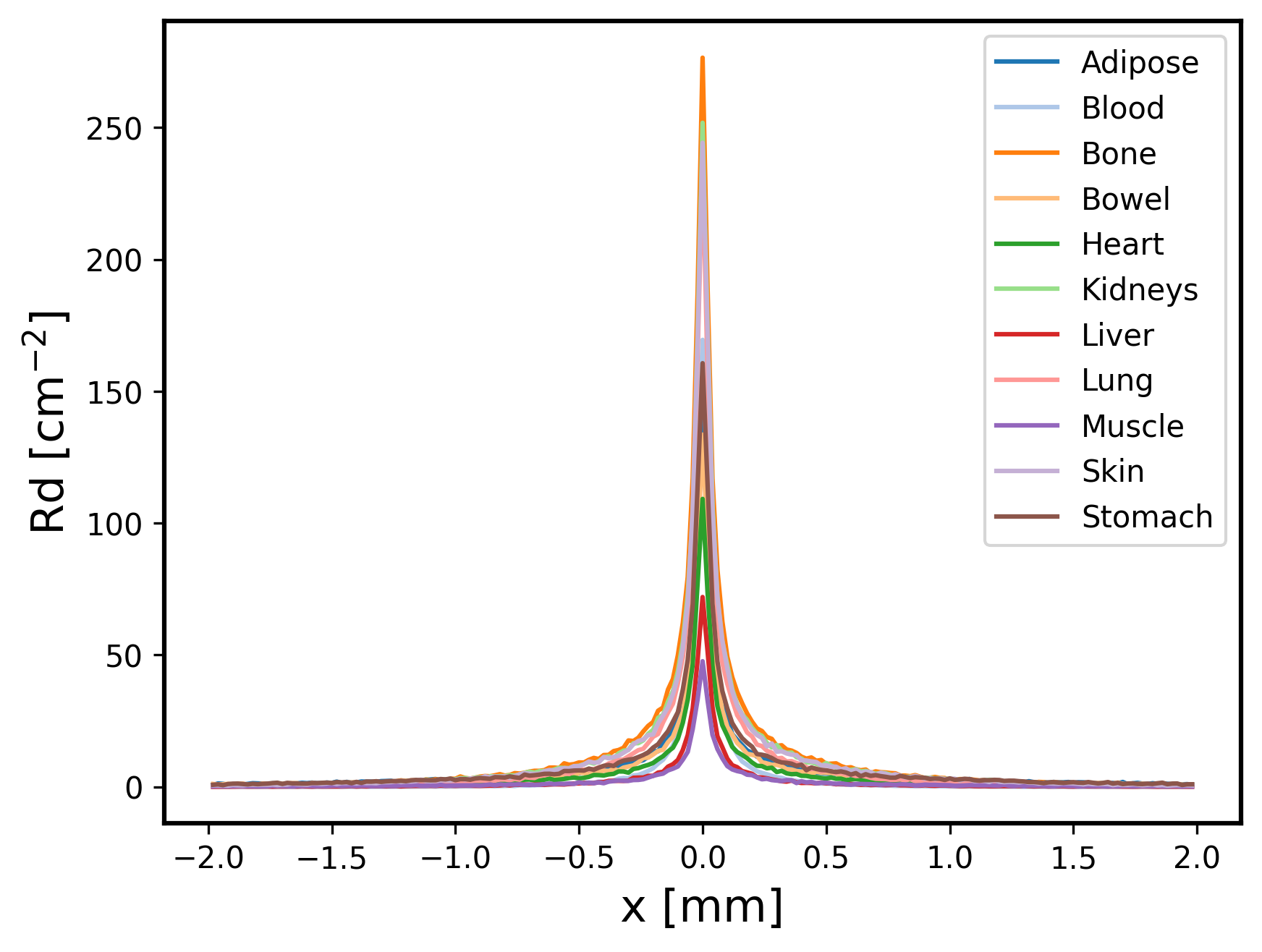}}
    \subfigure[]{\includegraphics[scale=0.52]{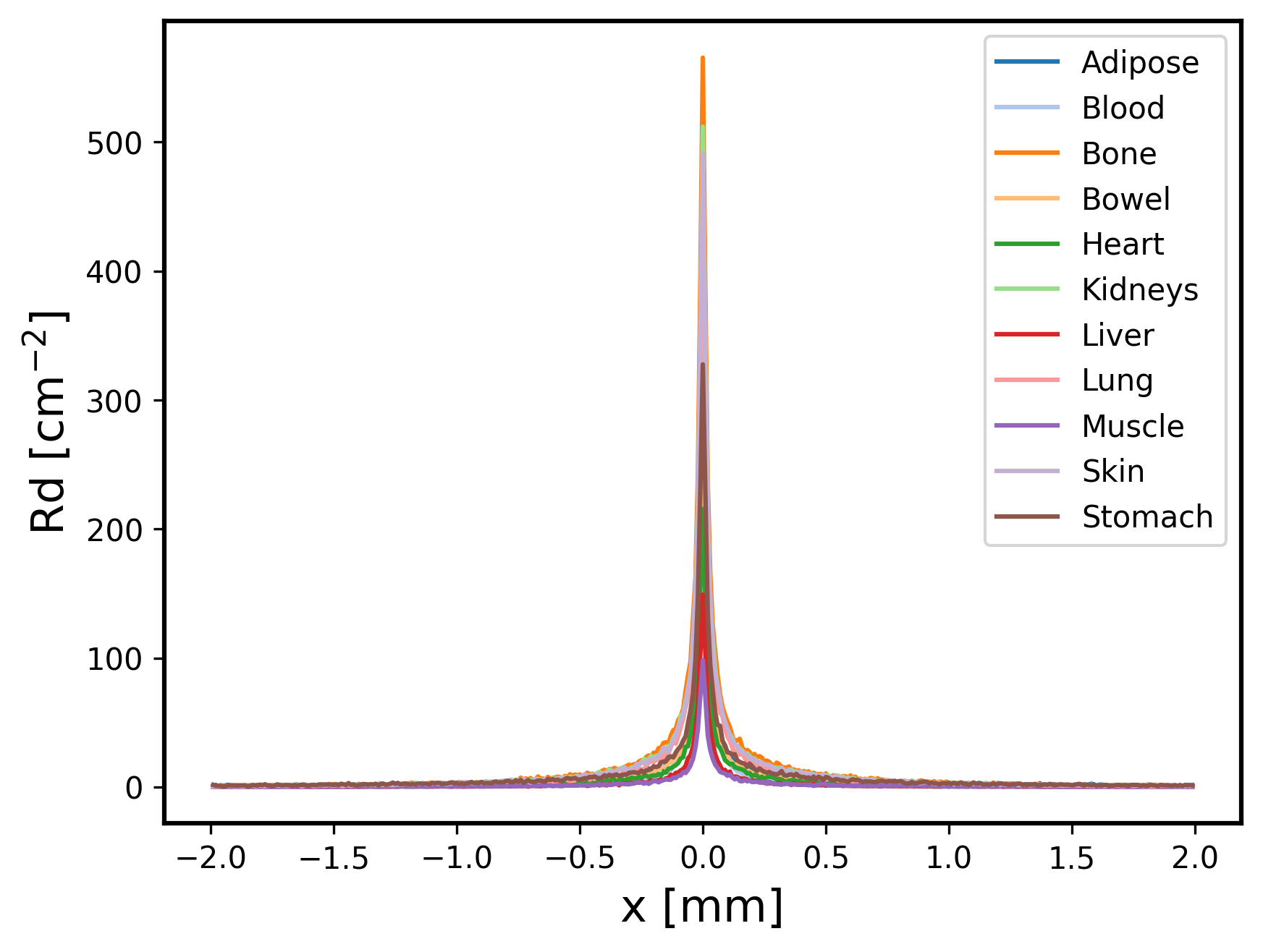}}
    \caption{Profiles through test images. (a) $DS_2$, FOV=4$\times$4@0.02, and (b) $DS_4$. FOV=4$\times$4@0.01. The anisotropy factor $g=0.6$ in both cases.  }
    \label{fig_profiles}
  \end{center}
\end{figure}

\subsection{Why Expanding FOV Unfavorable}

Considering the radial symmetry of the reflectance images, the horizontal profiles through the incident points of the test images in $DS_2$ and $DS_4$ are shown in Figure \ref{fig_profiles}. The profiles are extracted from the images with $g=0.6$ because the stronger scattering makes it easier to extract information on different tissues. 

Since our experimental setup employs an infinitely narrow incident beam and does not consider the camera model, the simulated reflectance images are actually the ideal impulse response of different tissues sampled at different FOVs and spatial resolutions. Figure \ref{fig_profiles}(a) shows the profiles at the FOV of 4$\times$4 and the resolution of 0.02. It can be seen that the reflectance signals are mainly concentrated in the range of [-1, 1] mm, corresponding to a FOV of 2$\times$2 mm$^2$. The estimation errors are comparable for the FOVs of 2$\times$2 and 4$\times$4 (Figure \ref{fig_accuracy_FOV}), with the errors for FOV 4$\times$4 a little better on average (Table \ref{tab_overview}). This indicates that the region between FOV 2$\times$2 and 4$\times$4 contains some discriminative information despite the difference is small between the reflectance signals. However, further increasing the FOV to 6$\times$6 would increase the estimation error significantly, as shown in Figure \ref{fig_accuracy_FOV} and Table \ref{tab_overview}. We believe that as the FOV expands, the reflectance signals become less distinguishable in terms of amplitude and distribution. 


As pointed out in \cite{2016_JBO_Quantifying}, the further away photons are detected from the incident beam, the deeper they have penetrated into the tissues. The reflectance intensities far away from the incident point are mostly contributed by the diffusive background, while the proximal intensities are composed of both the subdiffusive scattering and diffusive background. Since the diffusive background is weakly influenced by the scattering phase function, a more elaborate network design and more training samples may be required to take full advantage of it for phase function estimation.


\subsection{Why Refining Resolution Unfavorable}

Figure \ref{fig_profiles}(b) shows the profiles for the FOV of 4$\times$4 and the resolution of 0.01. Compared with \ref{fig_profiles}(a), the profiles have narrower lobes, larger amplitudes, and more clearly delineated fluctuations. This result is easy to interpret, since a higher spatial resolution is in use to sample the ideal impulse response. However, the distances between these tissue profiles have become smaller and even intersected, thus greatly increasing the difficulty for PhaseNet to discriminate different phase functions, compromising the estimation accuracy. 

The increased fine details in the reflectance images with high spatial resolution suggest that the learning task becomes more difficult. Without changing the network structure, a simple solution is to increase the number of training samples. To test this idea, the number of training samples per parametric setting in $DS_4$ was increased from 200 to 400, thus doubling the total number of training data. Five PhaseNet models were trained with the same hyperparameters as before and tested on the original test dataset $DS_4$. The resulting MSE is 0.016$\pm$0.017 and the relative error of $g$ is 4.201$\pm$3.805 \%. That is, the MSE has been decreased by 44.8\% with the doubled size of training data.




\subsection{Why Large $g$ Troublesome}

The results in Section \ref{section_results} show that at $g$=0.9 (the largest value we assumed), the estimation errors are significantly higher than that for other $g$ values, regardless FOV, resolution or tissue type. When the anisotropy factor is large, the propagation of light is dominated by forward-scattering. As a result, the reflectance image becomes much closer to a two-dimensional impulse function than that for a smaller $g$. This will require a higher spatial resolution to delineate the reflectance distribution. In other words, the reflection image at a large $g$ value is more likely to be spatially undersampled. 

\subsection{Why Low Scattering Tissues Troublesome}

Figure \ref{fig_tissue_accuracy_g_FOV} and \ref{fig_tissue_accuracy_g_Res} show that in most cases the estimation errors are larger for muscle, liver and blood than for the other tissue types. All the three types of tissues are low scattering, as classified in Table \ref{tab_tissueParam}. Low scattering means that more photons are forward scattered or absorbed. This results in a lower intensity and smaller spot size in reflectance images. As shown in Figure \ref{fig_profiles}, the lowest amplitude and the smallest width of the profiles are found for muscle and liver. Similar to the case of large $g$ values, the small main lobe width in the reflectance image explains why the estimation error becomes larger in the case of low scattering tissues.

\begin{figure}[!htbp]
  \begin{center}
    \subfigure[]{\includegraphics[scale=0.55]{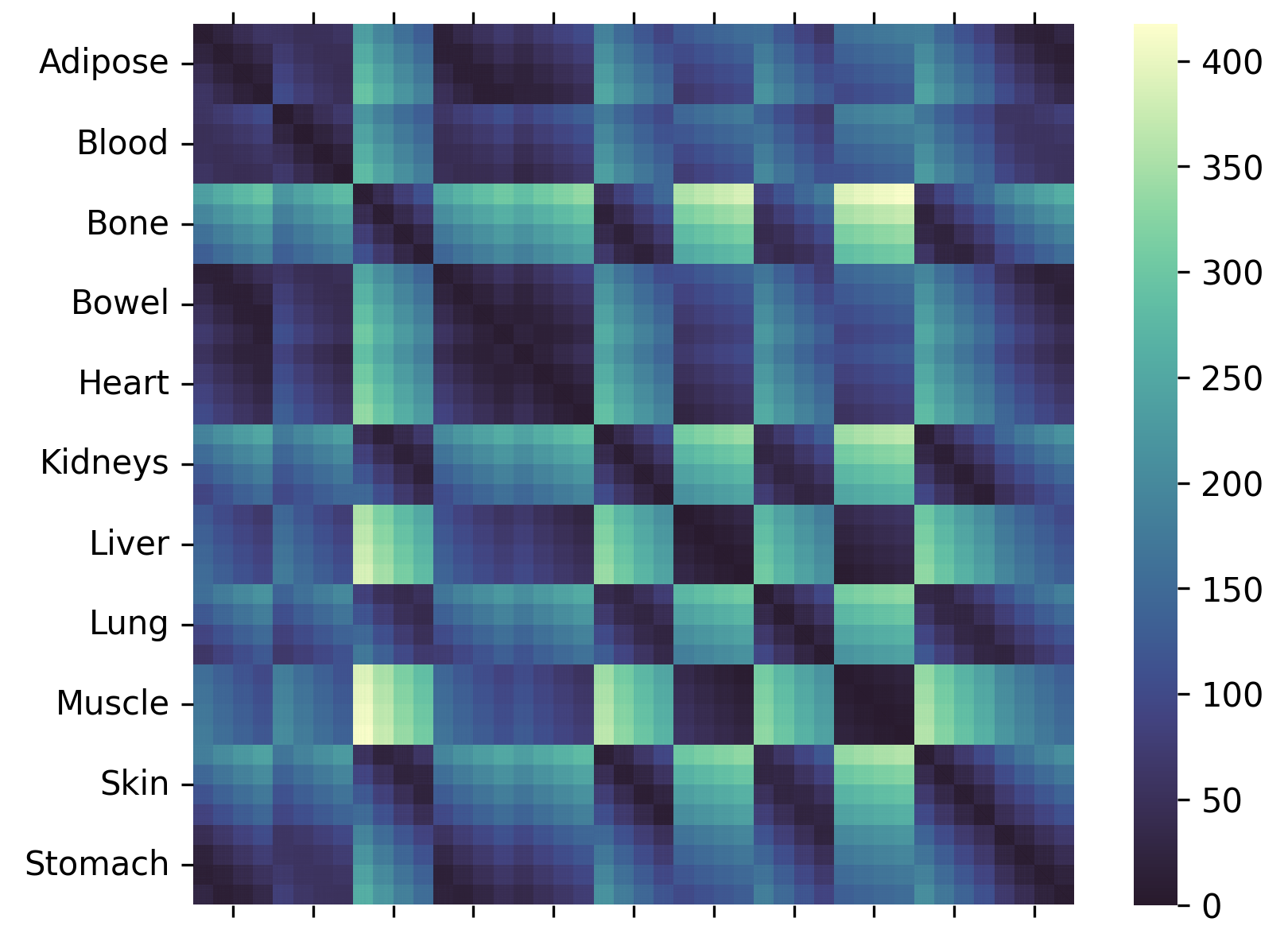}}
    \subfigure[]{\includegraphics[scale=0.55]{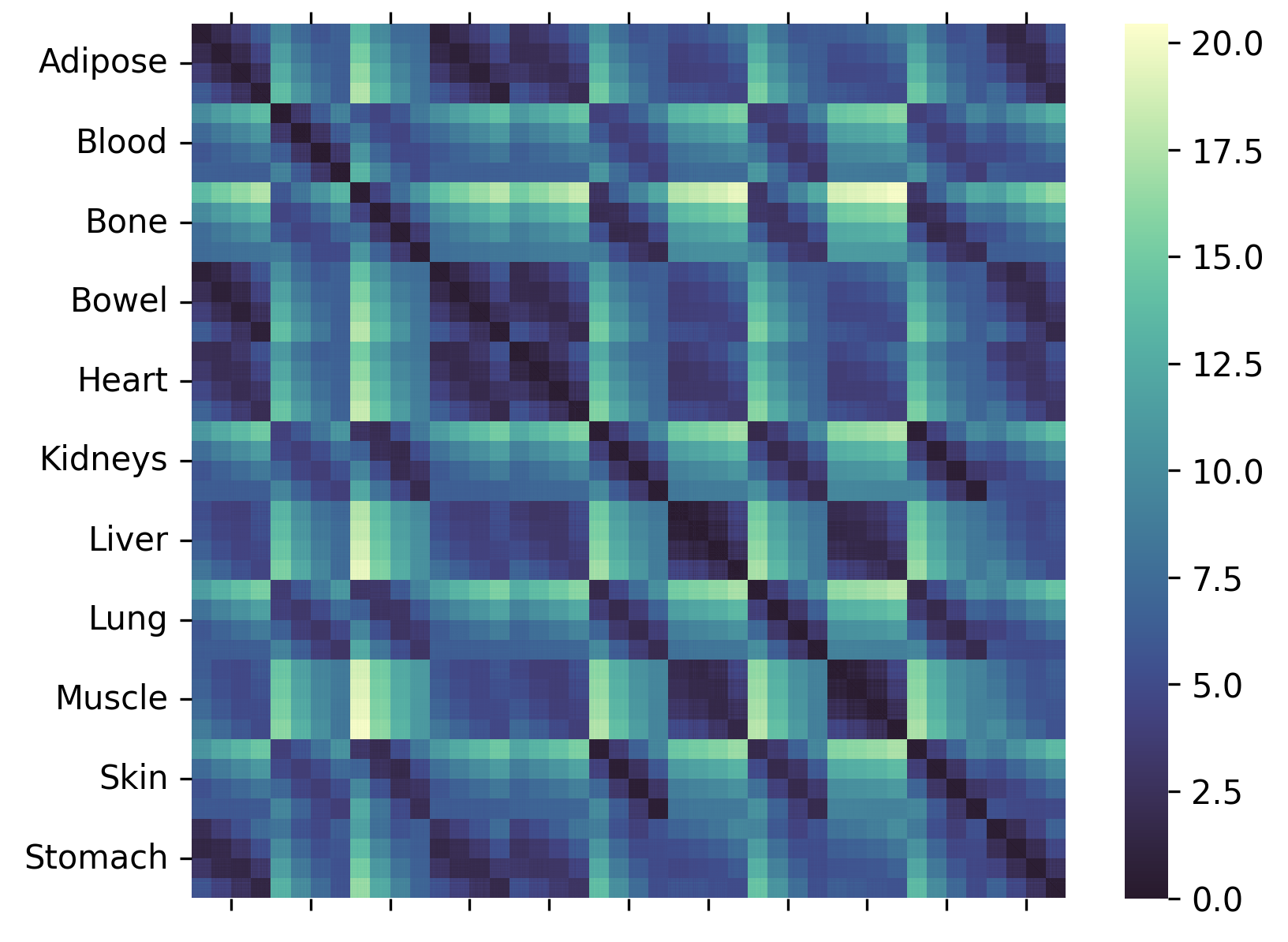}}
    \caption{Representational dissimilarity matrices (RDMs) of the test images in $DS_2$. (a) RDM of the profiles of the original images and (b) RDM of the image features.}
    \label{fig_RDM}
  \end{center}
\end{figure}

The representational similarity analysis (RSA) \citep{2008_RSA} can be employed to analyze this issue in more depth. RSA is widely used in neuroscience to compare representations across domains using the representational dissimilarity matrices (RDMs). RDMs are square symmetric matrices with zero diagonal that encode the (dis)similarity between all pairs of data samples or conditions in a dataset. We compute the RDMs using Euclidean distance for two representations of the reflectance image, namely the profiles of the original image and the image features extracted by PhaseNet. The output of the global pooling layer in PhaseNet is used as the image feature. The resulting RDMs of $DS_2$ are shown in Figure \ref{fig_RDM}. Since there are 1,760 images (11 tissues, 4 $g$ values, 40 samples per $g$ value) in the test set, the size of RDMs is 1,760$\times$1,760 pixels. The images are sorted by tissue type first and then by $g$ value. 

The RDMs in Figure \ref{fig_RDM} seem composed of small cells. The size of the cells is 40$\times$40 pixels, demonstrating that the variation between images generated with same parameters ($\mu_a$, $\mu_s$ and $g$) is small. A square of 4$\times$4 cells indicates the degree to which each pair of tissues is distinguished, or the differentiability of the same tissue with different $g$ values when the square is on the diagonal. It can be seen that there are four large dark squares of size 8$\times$8 cells in the top-left quarter of Figure \ref{fig_RDM}(a), indicating that adipose, blood, bowel and heart are closer to each other. The closer relationship between bone, kidneys, lung and skin can also be observed. It is worth noting that the four dark squares correspond to liver and muscle. They show not only that liver and muscle have similar reflectance characteristics but also that it is difficult to distinguish between livers or muscles with different $g$ values due to the homogeneity in the squares. Therefore, it is easy to understand why the estimation errors for liver and muscle are larger than that for other tissues. As for blood, it has the largest absorption coefficient among the tissues. Since a large amount of light is absorbed, the estimation error is large for high absorption tissues. Our results are consistent with the optical parameter estimation data in \citep{2021_SR}. In the RDM of the image features (Figure \ref{fig_RDM}(b)), the dark square effect is significantly weakened and the heterogeneity of each square is remarkably increased. This shows a better discriminability among involved tissue types and $g$ values in the feature space. 

\subsection{Limitations}

There are several limitations that should be acknowledged when interpreting the results of this study. First, the experimental configuration does not consider the incident light beam form, imaging system model, and noise distribution. The reflectance images are the ideal impulse response of tissues. Therefore, the FOV and resolution values in the experimental results are not applicable to a real imaging system. Second, the optimal number of Gaussian components for $DS_2$ is assumed for all other datasets because we hope to keep the experimental factors consistent except for FOV and resolution. This may degrade the performance on these datasets. Third, since FOV, resolution and image size are interrelated, the image size of the datasets should be adjusted as well after varying the FOV and resolution. Comparing the PhaseNet models with different input image sizes is not totally fair, but our results should be reasonable to show the feasibility, since we do not want to introduce interpolation errors by normalizing the image sizes.

\section{Conclusion}

In conclusion, we have proposed a novel deep learning approach to estimate the functional form and parameters of the phase function directly from diffuse reflectance data. The Gaussian mixture model has been used to specify the phase function using a modified ResNet-18 regression model. The proposed network  has been validated on MC simulated reflectance images of 11 biological tissues and the Henyey-Greenstein  phase function with typical anisotropy factors. In our experiments, the mean-squared-error and the relative error of the anisotropy factor are 0.01 and 3.28\% respectively, demonstrating the feasibility and accuracy of using a convolutional neural network for phase function estimation. A comparative study has been performed to analyze the effects of FOV and spatial resolution on the estimation accuracy, and provide the guidelines to optimize the experimental protocol.

\section*{Acknowledgments}

The authors at Rensselaer Polytechnic Institute would like to acknowledge the funding support from R01EB026646, R01CA233888, R01CA237267, R01HL151561, R21CA264772, and R01EB031102. The authors at Xidian University are supported by the National Natural Science Foundation of China under Grant No. 61976167 and the Key Research and Development Program in the Shaanxi Province of China under Grant No. 2021GY-082. 

\appendix



\bibliographystyle{dcu}
\bibliography{phase_reference}

\end{document}